Department of Computer Science
Georgia State University

**Master Project**

# Improved Detection of Adversarial Images Using Deep Neural Networks

*Student:*

Yutong Gao

*Directed by:*

Dr. Yi Pan

April 20, 2020

# *Abstract*

Machine learning techniques are immensely deployed in both industry and academy. Recent studies indicate that machine learning models used for classification tasks are vulnerable to adversarial examples, which limits the usage of applications in the fields with high precision requirements. We propose a new approach called Feature Map Denoising to detect the adversarial inputs and show the performance of detection on the mixed dataset consisting of adversarial examples generated by different attack algorithms, which can be used to associate with any pre-trained DNNs at a low cost. Wiener filter is also introduced as the denoise algorithm to the defense model, which can further improve performance. Experimental results indicate that good accuracy of detecting the adversarial examples can be achieved through our Feature Map Denoising algorithm.

# *I. Introduction*

Machine learning (ML) techniques are immensely employed in both industry and academy, achieving excellent performance on tasks such as classification, natural language processing, auto-driving, etc. Image classification, as one of the most popular domains, can be applied to many fields. The assumption underneath the image classification models is that training dataset share the indistinguishable data distribution with the testing dataset so that the learned features can be used to predict later. The consistency, however, can be broken by poisoning the training/testing dataset. Those poisoned examples, which are called adversarial examples, are generated by adding tiny perturbation into legitimate image in pixel value. An adversarial image usually is imperceptible for human at image level. The deviation between legitimate examples and adversarial examples will lower the performance of ML models. Samuel et al. [3] introduce real-world security issues related to the health information machine learning system with the intent of insurance claims approval. And Samuel suggests that defenses have to secure against all possible present and future attacks.

Several defense techniques [2] are proposed to mitigate the security issues caused by the "modified" examples. Some of the previous works aim to harden the ML models to be robust to the adversarial examples so that they can ensure the performance of the models (Section II), such as adversarial training and gradient masking. For adversarial training, it suffers from the costly model retraining. Apart from the expensive training cost, it also needs numerous adversarial examples which need to be generated for training. Gradient masking defense reduces the sensitivity of the model to small perturbation. Nevertheless, the less sensitive the model is, the less accuracy the model will have on predicting the legitimate images. A few other studies detect adversarial examples and leave the DNNs unchanged to avoid the high cost in the training stage (Section II). The reactive defense models can be used on distinct DNNs.

In this paper, we propose Feature Map Denoising defense that differentiates legitimate inputs and adversarial inputs generated by Fast Gradient Sign Method and Basic Iterative Method attack algorithms on ImageNet dataset. The underlying approach behind Feature Map Denoising is to compare the feature map of a given input, which can be represented by the prediction probability of the DNNs, with the feature map of the input after applying denoise techniques. In other words, the denoise method has different extents of effect on legitimate inputs and adversarial inputs. The degree of effects can be quantified by the distance between the prediction vectors of the denoised image and original one.

*Contribution*

In this work, we introduce a new method named Feature Map Denoising for detecting the adversarial examples, which can be used to associate with any pre-trained DNNs. Most of the previous studies perform the defense models on MNIST and CIFAR dataset, and the performance of the models will decrease when applied on more complex datasets. 1. We challenge to detect the adversarial inputs on one of the most complicated datasets, ImageNet. 2. To denoise the feature map, we apply the wiener filter, which is used to get involved as part of the defense model for the first time. 3. Our model can achieve a 67% overall detection rate of adversarial inputs generated by FGSM and BIM, which have even larger perturbation, on the ImageNet datasets.

## II. Related Works

In this section, we investigate and summarize the related works that connect with our problem.

Christian et al. [9] propose adversarial training to optimize the problem of the adversarial negatives. Adversarial examples with the ground-truth labels are added into the training dataset, so that the DNNs can learn the features of adversarial examples to predict correctly. Florian et al. [10] propose ensembled adversarial training. A larger scale of adversarial examples generated from multiple attack algorithms are used in the training stage. The methodology corporates the perturbation features which transfer from the holdout models to increase the robustness of the DNNs. However, generally, the adversarial learning needs the model to be trained on specific perturbation to gain robustness. The high cost of generating large scale of the adversarial datasets and retraining the model, especially on complicated models and datasets, is the weakness. Moreover, the precondition of training on the adversarial datasets is the attack algorithms should be known, which is not usually same in the real-world cases.

Papernot et al. [8] introduce the defense distillation to solve the security problems related to DNNs. The distilled knowledge extracted from DNNs transfers back to itself, which benefits from additional information encoded in class probabilities. In other words, the data distribution of the adversarial examples could be represented by the prediction vectors in another form, which can be used as distilled knowledge to be learnt by the DNNs. The model encapsulates distilled knowledge for the purpose of reducing the sensitivity, which can avoid the wrong output induced by small perturbation. However, the significant perturbation can be hardly defended by reducing the sensitivity, and the less sensitive the model is, the lower accuracy the model will has.

Xu et al. [13] propose feature squeezing to mitigate the security issues caused by malicious inputs. Instead of modifying the structure of the DNNs to increase the robustness, Xu's approach reduces the dimensions of the feature spaces of the inputs and leaves the DNNs unchanged. The intuitive reasoning is that unnecessary input spaces increase the chance of generating the perturbation that will confuse the prediction. The distance between the prediction vectors of the original input and squeezed input is calculated to be used as metrics to filter the malicious data. The threshold is selected to filter out the adversarial inputs linearly. The encouraging result shows on an even more complicated dataset, like ImageNet. However, the feature squeezing does not achieve ideal detection accuracy on some attacks with larger perturbation.

# III. Methods

### A. Fast Gradient Sign Method

Fast Gradient Signed Method is an algorithm described by Goodfellow et al. [4], which computes the perturbation to add on by utilizing the sign of the gradient of the objective function in aim to increase the loss of the DNNs. The formula can be expressed as:

$$X' = X + e * sign(\nabla J_x(X, y, \theta))$$

where e is the magnitude of perturbation, and $J(X, y, \theta)$ represents the loss of the neural network. The FGSM algorithm has significant superiority of computation time, and limits each pixel was changed up to the magnitude.

### B. Basic iteration method

Kurakin et al. [7] propose an extension attack algorithm, which applys the loss amplification in multiple iterations with small steps and limits the pixels change up to the *e* by clipping the intermediate result. The computation cost depends on the selection of the iterations. The direction of adding on pixel noises are transformed dynamically. The formula can be expressed as:

$$X'_0 = X, X'_{N+1} = Clip_{X, e}\{X'_N + esign(\nabla J_x(X'_N, y))\}$$

### C. Medium filter

The medium filter is one of the smoothing functions, in which to smooth the pixels by replacing the values of the neighbors. The sliding window runs over all signals and replaces each pixel by the median of its contiguous values within the window. The size of sliding window is the parameter and can be adjusted. The size of the window is positively related to the smoothness.

We run the medium filter on the images chosen from ImageNet, shown in *Figure 1*. The magnitude of the sliding windows is 3, which means every 3 x 3 window will be replaced by the median over 9 pixels. The usage of medium filter to denoise the feature maps intuitively decreases the effect of perturbation by smoothing the pixels, which drop the potential for DNNs to catch the adversarial pixels to alter the prediction. We assume that the difference between the prediction vector in the clean image and the denoised clean image is distinguishable from the difference of the adversarial image and the denoised adversarial images.

### D. Wiener filter

Norbert Wiener proposed wiener filter during the 1940s to mitigate the noise issue in the signals and the wiener filter has a good performance on noise reduction for several format of noises. The intention of the wiener filter is to find a filter g to estimate $\hat{x}$ which minimize

the mean square error with the unknown true $x$, the formula of wiener filter for denoising 2D signal is:

$$\hat{x}(m, n) = \left[\frac{H^*(m,n)}{|H(m,n)|^2+K}\right] y(m, n)$$

where $y\ (n, m)$ is the observed image and consists of true $x$ and unknown noise, $H\ (m, n)$ represents the degradation function, $H^*\ (m, n)$ represents the complex conjugate of $H\ (m, n)$, and K is constant. The mean square error is:

$$E = E|x\ (m, n) - \hat{x}\ (m, n)\ |^2$$

The images were transformed to grayscale since only two-dimensional wiener filter was formulated. And we run the wiener filter on the images chosen from ImageNet, and the examples shown in *Figure 2*.

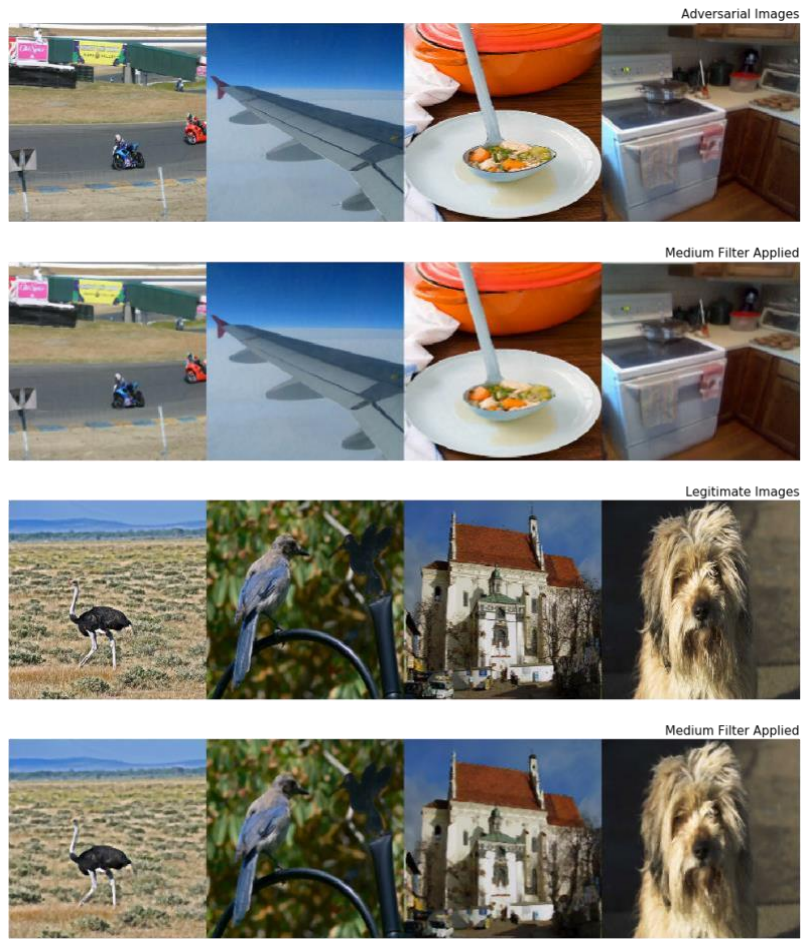

*Figure 1:* Medium Filter applied on legitimate images and adversarial images generated by FGSM attack algorithm. Adversarial images and denoised adversarial images shown in the first two rows; legitimate images and denoised legitimate images shown in the last two rows.

## E. ML classifiers

K Nearest Neighbor is a machine learning algorithm that takes the distance similarity as the measurement. During the training stage, the classifier stores all the available cases and corresponding labels. The test data are classified based on the similarity between data and previous training examples. In these experiments, the hyperparameter tuning was used and the K with the best detection accuracy during the training stage was selected.

The decision tree is a flowchart-like structure, as well as one of the supervised machine learning algorithms. Every node in the decision tree represents a specific test on single feature; every leaf node represents a class label, which is the decision made from the feature test. The strategy of the model is to split a dataset based on different conditions through both classification and regression tasks. The random forest classifier consists of multiple decision trees, which preserves the correct direction from the individual errors. The random forest performs better on the prediction that single tree has low correlation from others.

Support Vector Machine is a supervised learning algorithm, which is based on the idea of finding a hyperplane that separates the dataset into different domains. The kernel used in SVM is a way to map the original dataset into a high dimensional space to find the separator. Radial Basis Function is the kernel function used in the experiment.

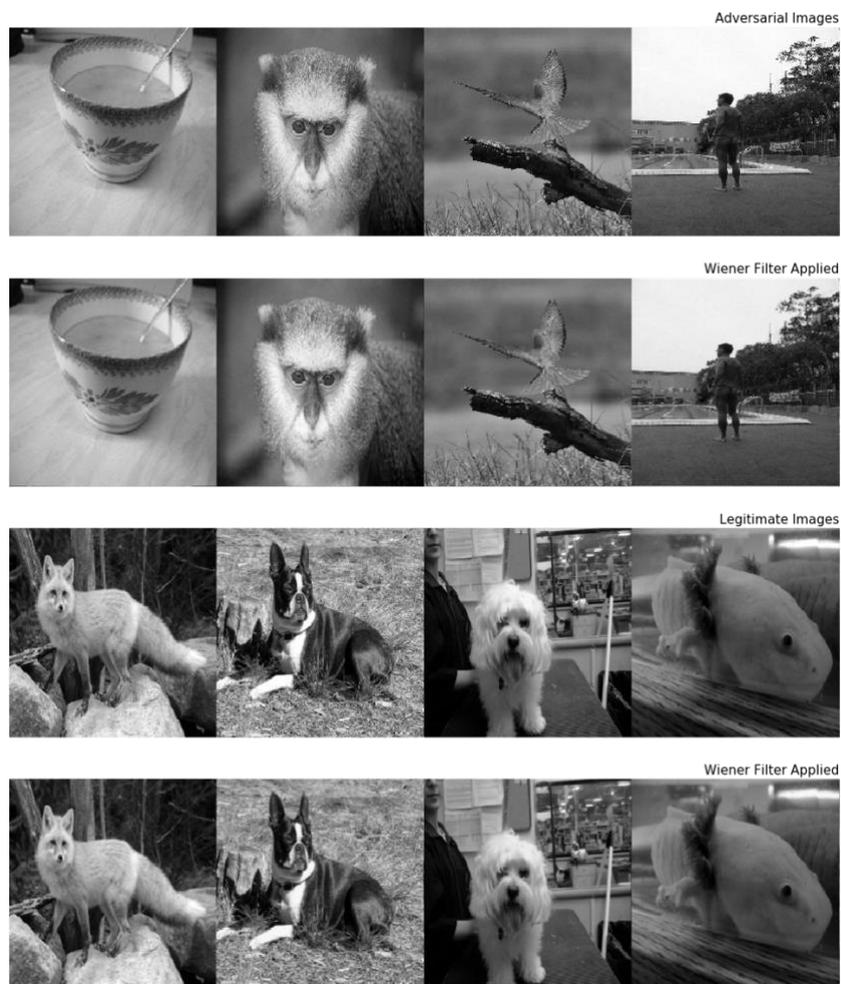

*Figure 2:* Winer Filter applied on legitimate images and adversarial images generated by BIM attack algorithm after transforming the images to grayscale. Adversarial images and denoised adversarial images shown in the first two rows; legitimate images and denoised legitimate images shown in the last two rows.

# IV. Experiments

In this section, we evaluate the performance of the Feature Map Denoising method to filter out the malicious inputs. We first test on adversarial examples generated by FGSM and BIM attack algorithms separately. Afterwards, we test on the mixed adversarial datasets to assess the performance of the defense model.

**Dataset**

ImageNet is a large-scale image dataset coordinated based on the WordNet hierarchy. With the development of the sophisticated algorithm on computer vision field, the ImageNet is inspired by the growing need of a uniform resource to test on. ImageNet contains over 20,000 synsets, and 1000 synsets with SIFT features. In this work, we randomly select the synsets and images in each synsets from 1000 classes with SIFT features.

Pre-generated adversarial examples from Xu's work [13]. Xu et al. select 100 images from ImageNet and apply the attack algorithms on to test the robustness. In our work, we use the same adversarial examples applied FGSM and BIM attack algorithms separately to test the performance of our model.

**Feature Map Denoise**

Medium filter and wiener filter are used to denoise the feature maps. We first test on the detection performance of the known attacks. Denoise method applies on the dataset which composed of the pre-generated adversarial images balanced with the legitimate images randomly selected from ImageNet separately. Then, we perform noise reduction on the dataset containing mixed adversarial images generated from two kinds of attack algorithms and clean images.

**Top-5 Confidence Prediction Scores**

The pre-trained MobileNet [5] predicts the original images and the denoise images separately. The output of the softmax layer which is one-dimensional prediction vector with the length of 1000 is transformed to the two-dimensional prediction vector with the length of 5 which consists of the top-5 confidence class labels and corresponding confidence scores. The average of the $L_0$ distance is computed between the top-5 confidence prediction vectors of the original image and denoise image, the formula of calculating the distance score is:

$$score(x) = \frac{1}{5} * \left\| \overrightarrow{P_5(x)} - \overrightarrow{P_5(x')} \right\|$$

where $\overrightarrow{P_5(x)}$ represent the top-5 confidence prediction vector of the original image, $P_5(x')$ represent the top-5 confidence prediction vector of the denoise image. (The distance is computed while the classes of the index are identical; if the class label does not exist in one of the vectors, the confidence score will set to be zero.

# ML classifiers

In this stage, multiple machine learning classifiers are used to classify the labeled scores. The labeled score has the format (X, y) where X represents the score computed from the previous step, and y ∈{1,0} represents adversarial or legitimate images.

The dataset was separated with a ratio of 0.5. We first test on the known attack with the k nearest neighbor, random forest, and decision tree. The hyperparameter tuning is performed during the training time and the best hyperparameters are used in the test stage. Support vector machine is used to classify the labeled score acquired from the wiener filter as well. Then, we test on the hybrid datasets, the hyperparameter tuning performed and the best classifier selected during the training stage, which used to test the performance of the detection. The overall approach of the Feature Map Denoising method is depicted in *Figure 3*.

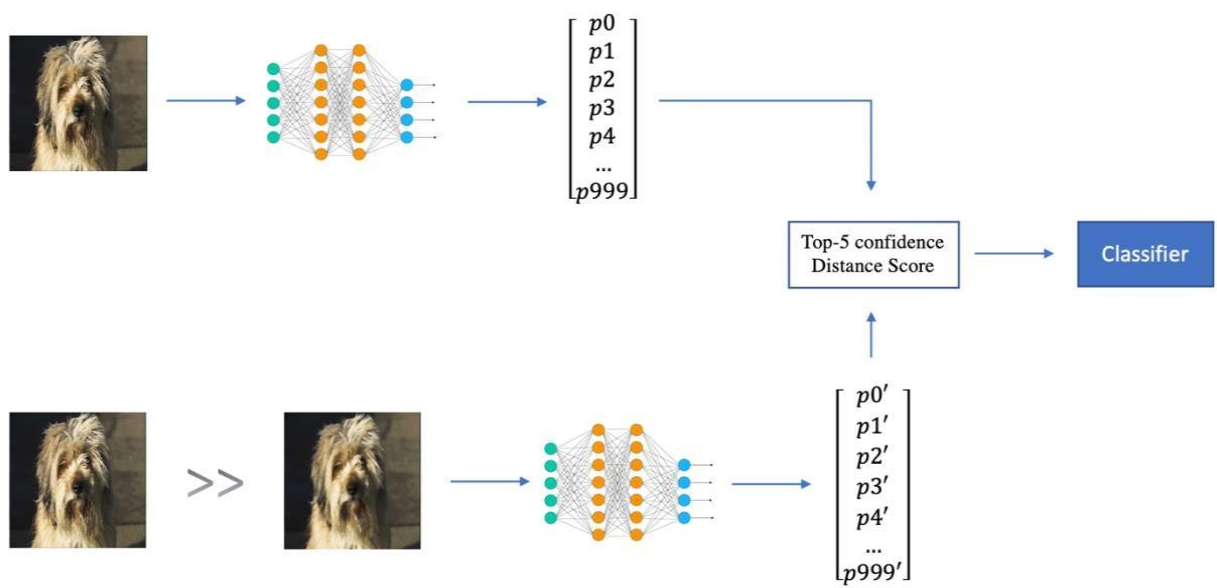

Figure 3. The flowchart of the Feature Map Denoising method.

## V. Results

The performance of the Feature Map Denoising defense is encouraging. The results of detection of the adversarial examples generated by single attack algorithms are shown in *Table 1*. The best accuracy of detecting the adversarial examples generated by FGSM attack algorithm is 68% with the medium filter; and 74% accuracy on detecting the adversarial examples generated by the BIM attack with the wiener filter. For the adversarial inputs generated by multiple attack algorithms, which is closer to the real-world cases, 67% overall accuracy can be achieved by applying the wiener filter denoise method. The comparison and individual detection rates are shown in *Table 2*.

|        | Medium Filter |      |      | Wiener Filter |      |      |      |
|--------|---------------|------|------|---------------|------|------|------|
|        | KNN           | DT   | RF   | KNN           | DT   | RF   | SVM  |
| FGSM   | 0.56          | **0.68** | 0.62 | 0.36      | 0.5  | 0.54 | 0.66 |
| BIM    | 0.4           | 0.42 | 0.48 | 0.56          | 0.58 | 0.56 | **0.74** |

Table 1. Result of adversarial examples detection rate of known attack algorithm.

|        | Xu et al. [13] | Medium Filter | Wiener Filter |
|--------|----------------|---------------|---------------|
| FGSM   | 0.4444         | 0.60          | **0.66**      |
| BIM    | 0.59           | 0.54          | **0.68**      |

Table 2. Result of adversarial examples detection rate of hybrid attack algorithm.

## VI. Discussion and Future Works

In this work, we propose the Feature Map Denoising algorithm which detects the hidden malicious images to increase the robustness of the machine learning models; and this method can be associate with any pre-trained models. Wiener filter, for the first time to be used to get involved as part of the defense model, makes essential contribution to the detection of malicious examples. It is not surprising that the defense model with the wiener filter performs better since the wiener filter denoises the image by minimizing the mean square error, which makes it closer to the clean image, whereas the medium filter only smooths the data distribution. By smoothing the adversarial image, the noise will still remain as both FGSM and BIM modify all the pixels of the images. To summarize, the defense model achieves an overall pleasing result on the complex dataset. Compared to the previous works, this work is inexpensively deployed, and it is efficient on the detection of the adversarial examples generated by FGSM and BIM.

Feature Map Denoising defense raises an early research on detecting the adversarial examples by classifying the distance scores between the probability vectors of the denoised and original input. To improve the current research, a wiener filter algorithm can be designed which can be applied on color images. While the approach provides pleasing results on defending some attack algorithms, it still needs to be extended to a universal scenario since unknown, mixed, and complicated attacks happen commonly in real-world cases. The ideal defense will secure against all possible present and future attacks. Moreover, 67% blended detection rate is encouraging but still not secure enough for some fields, like medical system. We plan to understand our prediction better through rule generation [6] and use other machine learning technologies such as Clustering SVM [16], Genetic Algorithms [17], [18], and Rough Sets [19], [20], [21]. Future research could investigate the change of feature maps before and after denoising by implementing deep learning, and some optimizing methods [1], [11], [12], [14], [15] and distributed methods [22] can be used to address the high cost of deep learning models.